\documentclass{article}

\usepackage[nonatbib, preprint]{style}

\usepackage[utf8]{inputenc} 
\usepackage[T1]{fontenc}    
\usepackage{hyperref}       
\usepackage{url}            
\usepackage{booktabs}       
\usepackage{amsfonts}       
\usepackage{nicefrac}       
\usepackage{microtype}      
\usepackage{subcaption}
\usepackage{graphicx}

\title{FAT ALBERT: Finding Answers in Large Texts using Semantic Similarity Attention Layer based on BERT}

\author{Omar Mossad\textsuperscript{1}, Amgad Ahmed\textsuperscript{2}, Anandharaju Raju\textsuperscript{3}, \\ Hari Karthikeyan\textsuperscript{4}, and 
        Zayed Ahmed\textsuperscript{5} ~\\[12pt]\normalsize 
~\fontsize{11}{13}\selectfont{Simon Fraser University\unskip, Burnaby\unskip, Canada}\\ \texttt{{
\{\textsuperscript{1}\href{mailto:omossad@sfu.ca}{omossad}, \textsuperscript{2}\href{mailto:amgada@sfu.ca}{amgada}, \textsuperscript{3}\href{mailto:aduraira@sfu.ca}{aduraira}, \textsuperscript{4}\href{mailto:hkarthik@sfu.ca}{hkarthik}, \textsuperscript{5}\href{mailto:zayed_ahmed@sfu.ca}{zayed\textunderscore ahmed}}\}@sfu.ca}}

\author{Omar Mossad\textsuperscript{1}, Amgad Ahmed\textsuperscript{2}, Anandharaju Raju\textsuperscript{3}, \AND Hari Karthikeyan\textsuperscript{4}, and Zayed Ahmed\textsuperscript{5} 
\\ \\
Simon Fraser University, \\ Burnaby, Canada \\ \\ 
\texttt{{
\{\textsuperscript{1}\href{mailto:omossad@sfu.ca}{omossad}, \textsuperscript{2}\href{mailto:amgada@sfu.ca}{amgada}, \textsuperscript{3}\href{mailto:aduraira@sfu.ca}{aduraira}, \textsuperscript{4}\href{mailto:hkarthik@sfu.ca}{hkarthik}, \textsuperscript{5}\href{mailto:zayed_ahmed@sfu.ca}{zayed\textunderscore ahmed}}\}@sfu.ca}}

\begin{document}

\maketitle

\begin{abstract}
Machine based text comprehension has always been a significant research field in natural language processing. Once a full understanding of the text context and semantics is achieved, a deep learning model can be trained to solve a large subset of tasks, e.g. text summarization, classification and question answering. \\
In this paper we focus on the question answering problem, specifically the multiple choice type of questions. We develop a model based on \textit{BERT}, a state-of-the-art transformer network. Moreover, we alleviate the ability of \textit{BERT} to support large text corpus by extracting the highest influence sentences through a semantic similarity model.
Evaluations of our proposed model\footnote[1]{\textbf{The source code} is publicly available https://github.com/omossad/fat-albert} demonstrate that it outperforms the leading models in the MovieQA challenge and we are currently \textbf{ranked first}\footnote[2]{\textbf{MovieQA Leader Board:} \href{http://movieqa.cs.toronto.edu/leaderboard/\#table-plot}{http://movieqa.cs.toronto.edu/leaderboard/\#table-plot}} in the leader board with test accuracy of 87.79\%. Finally, we discuss the model shortcomings and suggest possible improvements to overcome these limitations.
\end{abstract}

\section{Introduction}
One of the main challenges in Natural Language Processing (NLP) is the ability of a machine to read and understand an unstructured text and then reason about answering some related questions. Such question answering (QA) models can be applied to a wide range of applications such as financial reports, customer service, and health care.\par
A significant number of datasets namely \textit{MovieQA} \cite{movieqa}, \textit{RACE} \cite{race} and \textit{SWAG} \cite{swag} were created to provide a ground truth for training and evaluating a specific type of QA models that involve Multiple Choice Questions (MCQs). The \textit{MovieQA} dataset challenge has attracted a large number of promising solutions, such as Blohm et al. \cite{lstm}, where they implement two attention models based on Long-Short-Term-Memory (LSTM) and Convolutional Neural Networks (CNNs). Moreover, they combine both model accuracies using an ensemble aggregation. However, these models are prone to  various systematic adversarial attacks like linguistic level attacks (word vs. sentence level) and the knowledge of the adversaries (black-box vs. white-box). These models only learn to match patterns to select the right answer rather than performing plausible inferences as humans do. \par  

Recently, \cite{bert} implemented Bidirectional Encoder Representations from Transformers \textit{BERT}, which has since been used as a pre-trained model to tackle a large subset of NLP tasks. \textit{BERT}’s key technical innovation is applying the bidirectional training of Transformer, a popular attention model, to language modelling. The paper’s results show that a language model which is bidirectionally trained can have a deeper sense of language context and flow than single-direction language models. A plethora of deep learning models have ever since incorporated \textit{BERT} into several Tasks, ever-improving the state-of-the-art performances.\par
A notable limitation of \textit{BERT} is that it is not able to support large texts that include more than the pre-trained model's maximum number of words (tokens). Therefore, when dealing with large texts, the performance of \textit{BERT} is severely affected.
In our work, we aim to overcome the limitations of \textit{BERT} by analyzing and improving how accurate we predict the answer extracted from a large text in the \textit{MovieQA} MCQ dataset.
Our approach relies on the concept of \textit{sentence attention} to extract the most significant sentences from a large corpus. We were able to accomplish this task using a pre-trained \textit{semantic similarity model}.\par
The remainder of this paper is organized as follows: Section \ref{sec:datasets} describes the datasets focusing on the semantics and the nature of the questions. Next, we highlight the proposed model and the intuition behind all the components we used in section \ref{sec:model}. We evaluate the performance of our model in section \ref{sec:evaluation}. Finally, we conclude our report and suggest future modifications for the model in section \ref{sec:conc}.

\section{MovieQA Dataset}
\label{sec:datasets}
In this section, we give an overview of the dataset used to evaluate our model.
The MovieQA \cite{movieqa} dataset was created by generating a number of MCQs that can be solved from specific context extracted from the plots of real movies. An existing \textit{challenge}\footnote[1]{\textbf{MovieQA Challenge:} \href{http://movieqa.cs.toronto.edu/}{http://movieqa.cs.toronto.edu/}} is still on going to find the highest accuracy in solving these MCQs on movie topics. The models can be trained to use either video scenes, subtitles, scripts or movie plots (extracted from Wikipedia). The leader board for this challenge is divided according to the source of input selected for the model.
The dataset consists of almost \textit{15,000} MCQs obtained from over \textit{400} movies and features high semantic diversity.
Each question comes with a set of \textit{5} highly plausible answers; only \textit{one} of which is correct. 
The dataset structure and semantics for the movie plots are described in Table \ref{table:movieqa}.
On average, each movie plot has 35.2 sentences and there are 20.3 words per sentence on average. 
All training and validation sets are labelled with the correct answer. However, the test dataset is not labelled and can only be evaluated using the challenge's submission server. Due to the large nature of the plot texts, we have selected this dataset to demonstrate how we can incorporate BERT in relatively large texts.

\begin{table}
  \caption{MovieQA dataset description}
  \label{table:movieqa}
  \centering
  \begin{tabular}{lcccc}
    \toprule
    \multicolumn{1}{l}{\bf}  &\multicolumn{1}{c}{\bf Train} &\multicolumn{1}{c}{\bf Val.} &\multicolumn{1}{c}{\bf Test} &\multicolumn{1}{c}{\bf Total} 
    \\ \hline
    \# of Movies & 269 & 56 & 83 & 408 \\
    \# of Questions & 9848 & 1958 & 3138 & 14944\\
    Avg. Q. \# of words & 9.3 & 9.3 & 9.5 & 9.3 $\pm$ 3.5 \\
    Avg. CA. \# of words & 5.7 & 5.4 & 5.4 & 5.6 $\pm$ 4.1 \\
    Avg. WA. \# of words & 5.2 & 5.0 & 5.1 & 5.1 $\pm$ 3.9 \\
    \hline
    \bottomrule
  \end{tabular}
\end{table}

\section{FAT ALBERT Model Description}
\label{sec:model}
 Existing \textit{BERT} MCQ codes currently lack the support for large text documents since they are restricted to sequence length of at most 512 tokens. Furthermore, due to our limited compute capabilities we are restricted to $\leq$ 130 tokens. According to \cite{bert}, only TPUs with 64GB memory are able to train models with 512 number of tokens, compared to 16GB GPUs.\par
Therefore, we have used another model on top of \textit{BERT} MCQ to select the top 5 sentences from the text that are highly similar to the question context. Subsequently, instead of processing the entire corpus, \textit{BERT} MCQ uses the top 5 sentences only. 
\begin{figure*}[!htb]
\begin{center}
\framebox{\includegraphics[width=0.99\linewidth]{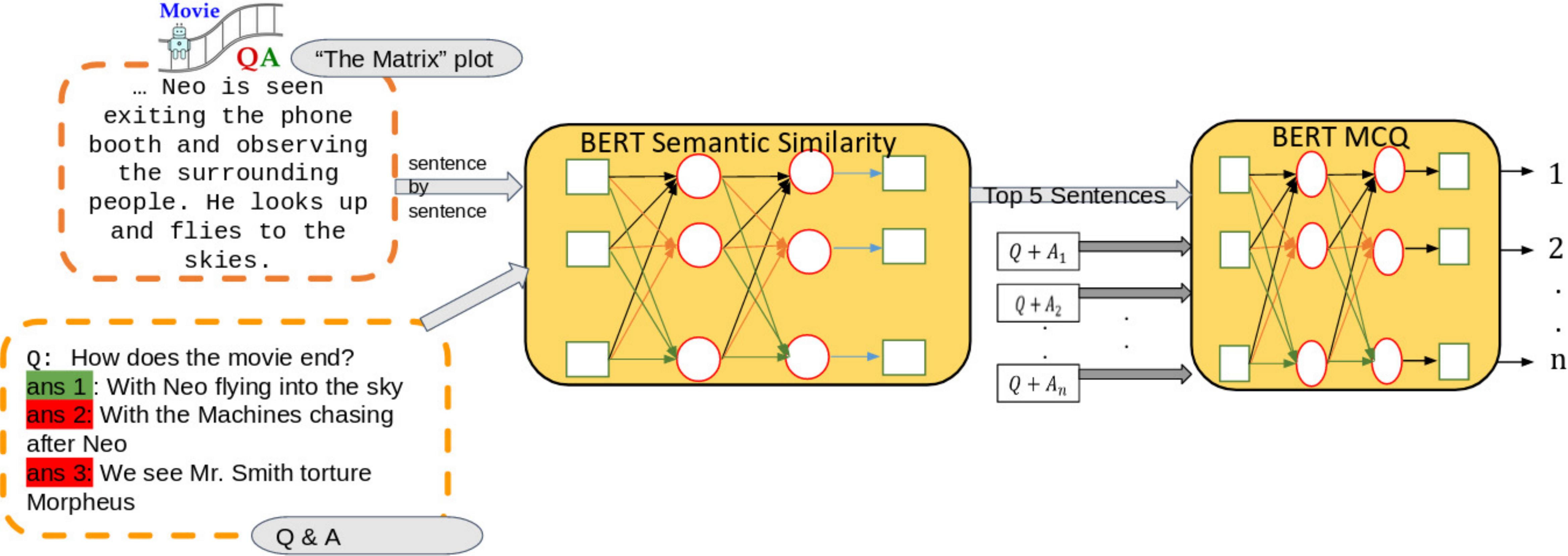}}
\end{center}
\caption{FAT ALBERT model overview}
\label{fig:model}
\end{figure*}
The maximum span of a certain question (i.e. the part from the text needed to answer this question)  is 5 sentences, and all plot alignments are consecutive (i.e. all questions span a specific passage from the text, not the entire text).\par
Our model consists of the following:
\begin{itemize}
    \item Semantic Similarity Classifier pre-trained on STS and Clinical datasets
    \item \textit{BERT} for MCQ trained on MovieQA dataset
\end{itemize}
A description of the entire model is depicted in Fig.\ref{fig:model}. The large movie plot text along with the concatenated question and answer sequences are fed to the similarity model to produce the top 5 similar sentences for each question.
We feed the questions, answers and the attention made plot text (top 5 similar sentences) to the \textit{BERT} for MCQ model. The output of this model is an array of probabilities having the size of the number of possible choices. Finally, we select the choice having the highest probability.

\subsection{Semantic Similarity Network}
\begin{figure}[t]
\begin{center}
\includegraphics[width=0.7\linewidth]{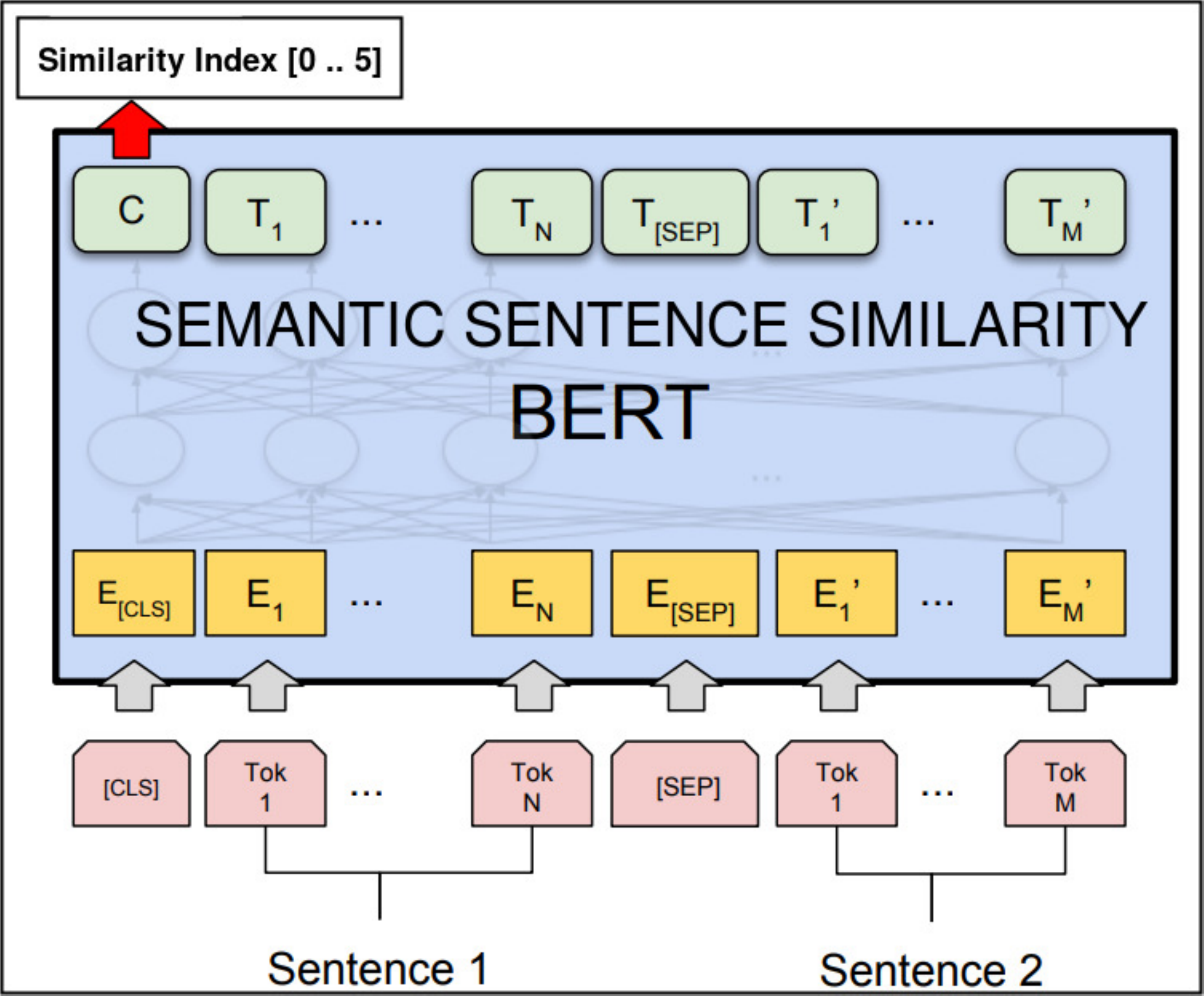}
\end{center}
\caption{Semantic Similarity Network model}
\label{fig:sim}
\end{figure}
We use two pre-trained models based on STS and Clinical datasets to find the similarity measure between 2 sentences. A number of variation for semantic similarity exists between our dataset and the aforementioned ones, but the selected model has proven to be effective in our application. A detailed comparison between the performance of these models on the MovieQA dataset is provided in the evaluation section.
The model uses \textit{BERT} to extract the embeddings from both sentences. These embeddings undergo a number of similarity functions, namely \textit{cosine similarity, qgram distance, levenshtein similarity,} etc., and the outputs are sent to a fully connected network followed by a softmax layer to provide the similarity index. We have selected to use a combination of STS and Clinical \textit{BERT} models after normalizing each model probabilities. Eq. \ref{eqn:cos} describes how the cosine similarity index between two sentences $s_{1}$ and $s_{2}$ is calculated.
Fig. \ref{fig:sim} highlights the contextual structure of the \textit{BERT} similarity model.
\begin{equation}
\cos ({\bf s_{1}},{\bf s_{2}})= {{\bf s_{1}} {\bf s_{2}} \over \|{\bf s_{1}}\| \|{\bf s_{2}}\|} = \frac{ \sum_{i=1}^{n}{{\bf s_{1}}_i{\bf s_{2}}_i} }{ \sqrt{\sum_{i=1}^{n}{({\bf s_{1}}_i)^2}} \sqrt{\sum_{i=1}^{n}{({\bf s_{2}}_i)^2}} }
\label{eqn:cos}
\end{equation}

\subsection{BERT for MCQ}
The \textit{BERT} MCQ model uses a pre-trained BERT transformer network modified and fine tuned on the MovieQA dataset.
The embedding outputs of \textit{BERT} are passed to a fully connected layer to produce the predicted probabilities of each possible choice. 
We used the \textit{bert\textunderscore large\textunderscore uncased} pre-trained model which uses 24-layer, 1024-hidden, 16-heads and 340M parameters. The fine tuning is performed on the MovieQA dataset after modifying \textit{BERT} outputs to support the variable number of choices.
When running \textit{BERT} for MCQ on the perfectly aligned plot sentences, the model was able to achieve a validation accuracy of 92.34\%. Although, the model was initially developed to support sentence completion type of questions, we modified the model to handle MCQs by changing the network structure, to output probabilities for each choice instead of complete text tokens.

\section{Evaluation Results}
\label{sec:evaluation}
We evaluate the performance of our model on the MovieQA dataset. In this section, we indicate whether the results were obtained from our own implementation or as mentioned in the reference paper. Some differences appear between our evaluation and the published results, probably due to changes in parameters by the authors which were not mirrored in their source codes. We also provide a brief case study to highlight two cases from the same movie plot where the model succeeds and fails, respectively.
\subsection{Evaluation on MovieQA Dataset}
A properly trained \textit{BERT} for MCQ can reach an accuracy in the range of 90\% on the MovieQA dataset once it is aligned with exact sentences having the answer. However, when performed without any sentence selection, the accuracy drops to 20\% (a random choice) due to the fact that \textit{BERT} truncates the sentences with word count higher than the maximum sequence length (130 words in our case).
Therefore using the semantic similarity model we captured the top 5 similar sentences to the question at hand. We selected 5 questions since the average word count of 5 sentences over the entire dataset is in the range of 110 which seems to be acceptable to truncate a few words from the least similar sentences. The similarity model performance is depicted in table \ref{table:sim}. The combined model aggregates the output probabilities of Web and Clinical models after normalization. A possible improvement for the model accuracy can be done by further training it on the MovieQA dataset.

Next, we compare the accuracy of our model against the current MovieQA challenge leader board models. We have included the validation accuracy and test evaluations we received from the MovieQA authors after submission.
Another contribution, is that we have created an ensemble model that aggregates the results from the top 4 approaches in the challenge and performs a majority ruling to select the label with the highest probability. This model uses the CNN and LSTM models previously described along with the \textit{BERT} model. The main incentive behind this ensemble is to allow different models to correct one another and collaboratively avoid making mistakes. Table \ref{table:movieqa_eval} demonstrates the accuracy of our models compared to the leader board models.

\begin{table}
  \caption{MovieQA semantic similarity evaluations}
  \label{table:sim}
  \centering
  \begin{tabular}{l c c}
    \toprule
    \multicolumn{1}{l}{\bf}  &\multicolumn{1}{c}{\bf Train. Acc.\%} &\multicolumn{1}{c}{\bf Val Acc.\%}
    \\ \hline
    Web BERT & 88.87 & 89.03 \\
    Clinical BERT & 88.12  & 89.28  \\ 
    Combined & 90.27 & 91.07
    \\ \hline
    \bottomrule
  \end{tabular}
\end{table}

\begin{table}
  \caption{MovieQA dataset evaluations}
  \label{table:movieqa_eval}
  \centering
  \begin{tabular}{l c c}
    \toprule
    \multicolumn{1}{l}{\bf}  &\multicolumn{1}{c}{\bf Val. Acc.\%} &\multicolumn{1}{c}{\bf Test Acc.\%}
    \\ \hline 
    FAT ALBERT (Ensemble) & \textbf{87.48} &   \\
    FAT ALBERT & 87.28 & \textbf{87.79} \\
    LSTM \cite{lstm} & 80.90 / 83.14* & 85.12 \\
    CNN \cite{lstm} & 79.47 / 79.62* & 82.73 \\
    Word-Level CNN \cite{lstm} & 73.39 / 76.50* &
    \\ \hline
    \bottomrule
  \end{tabular}
    \begin{center}
    \small * as included in the paper/leader board (not obtained from our evaluations)
    \end{center}
\end{table}

\begin{figure}[t]
        \begin{subfigure}{0.5\textwidth}
                \centering
                \fbox{\includegraphics[width=0.99\linewidth]{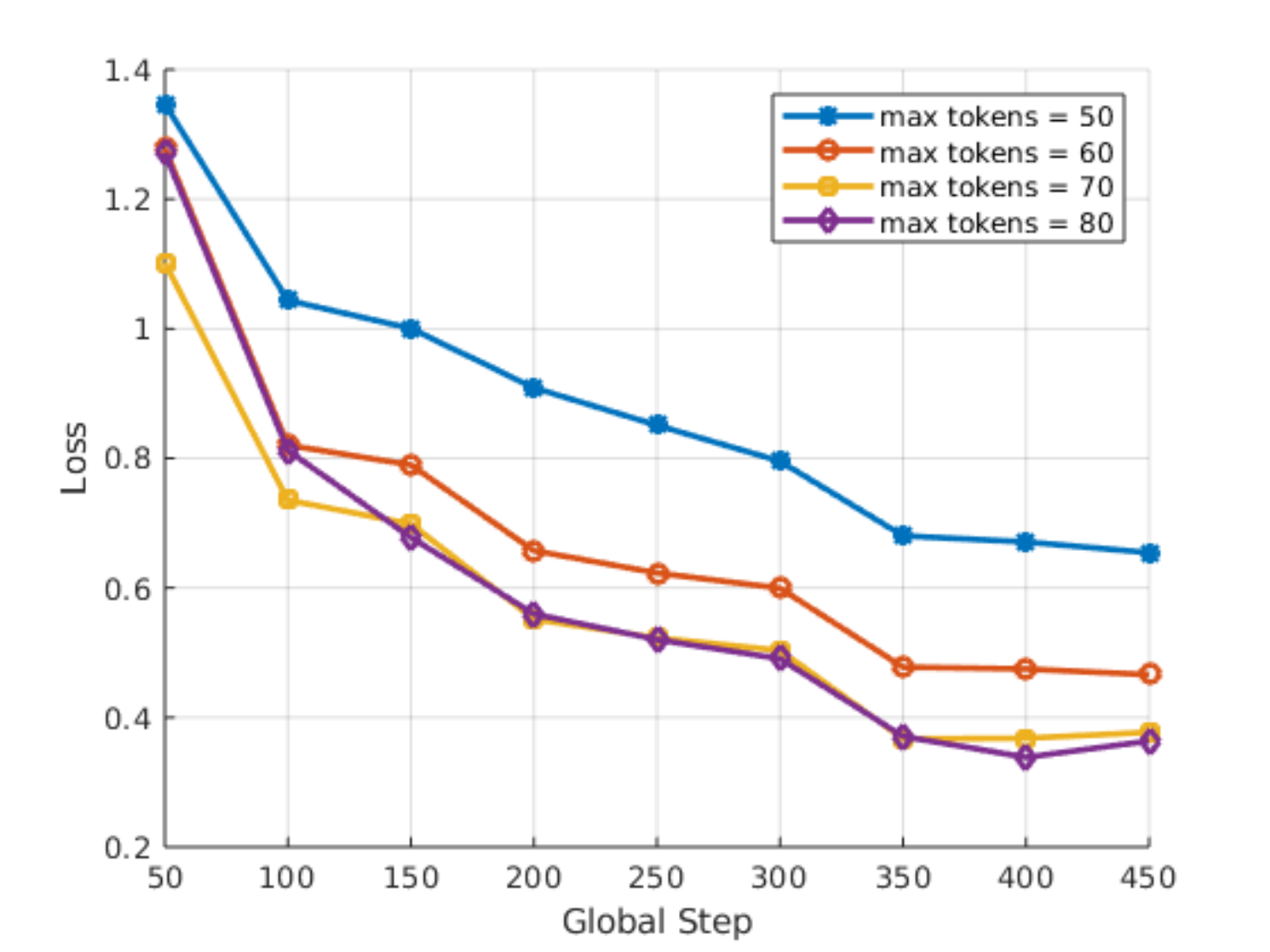}}
                \caption{Batch size = 8}
                \label{fig:loss_8}
        \end{subfigure}%
        \hspace{15pt}
        \begin{subfigure}{0.5\textwidth}
                \centering
                \fbox{\includegraphics[width=0.99\linewidth]{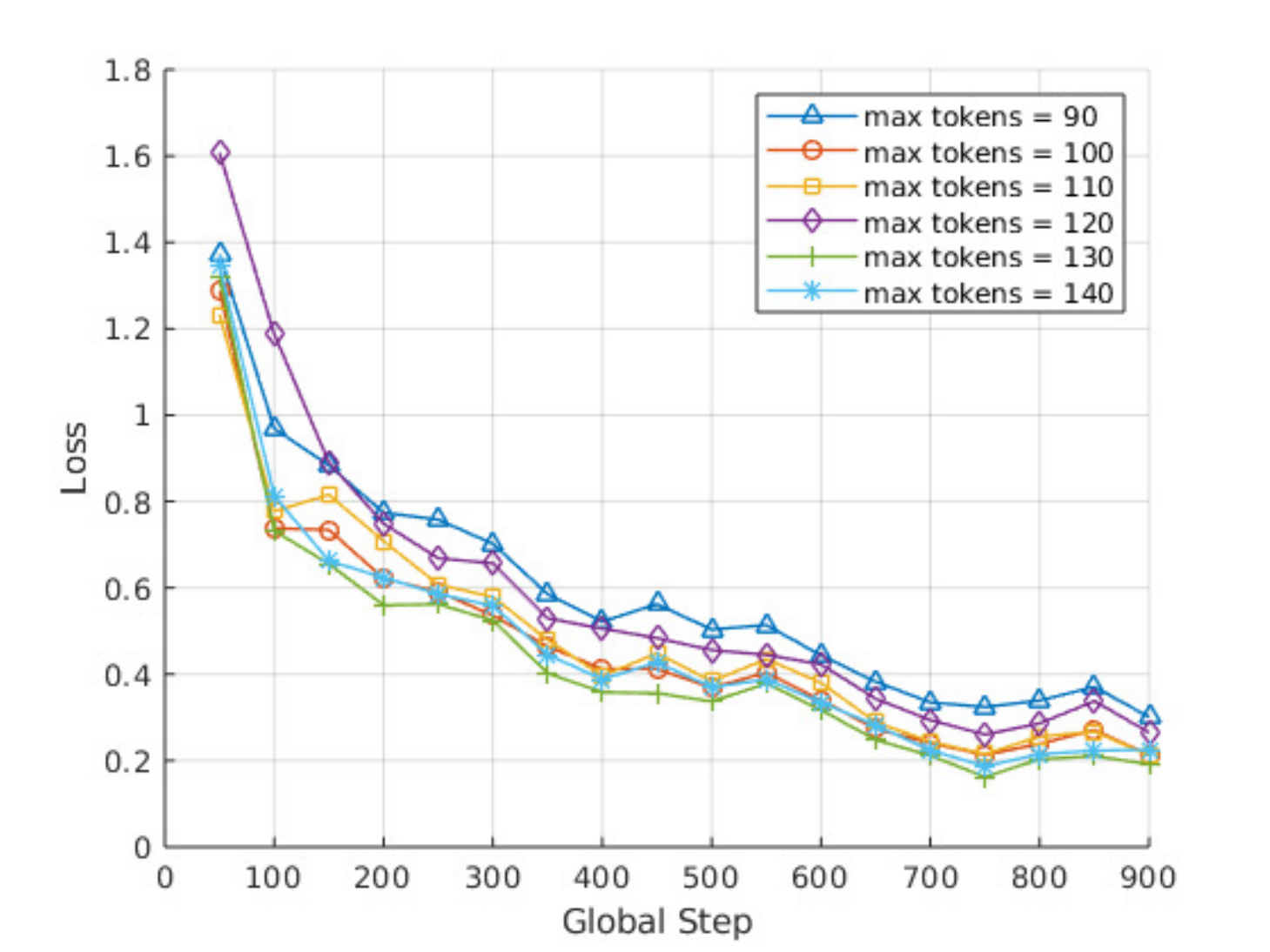}}
                \caption{Batch size = 4}
                \label{fig:loss_4}
        \end{subfigure}%
        \caption{BERT MCQ training loss on MovieQA dataset}
        \label{fig:loss}

\end{figure}

To highlight the effect of the number of tokens in BERT, we showcase in Fig. \ref{fig:loss} the training loss for models with different number of tokens. The main observation is that in order to support larger number of tokens, we had to reduce the batch size used during training. Although, the results indicate that larger tokens have lower losses in general, it is clear that reducing the batch size has notably affected the model accuracy. For instance, in Fig. \ref{fig:loss_8}, when using a number of tokens equivalent to 80, the model accuracy increases significantly compared to a maximum number of 50 tokens where the model truncates any input having $\geq 50$ tokens. Due to the compute capability we haven't been able to run the model with $\geq 140$ tokens as the GPUs were not able to allocate enough memory. Therefore, we evaluated the loss for larger number tokens using a smaller batch size in Fig. \ref{fig:loss_4}. The loss is gradually decreasing as the number of tokens becomes higher until it stabilizes when the inputs generally become smaller than the maximum number of tokens. Hence, the input sentences are padded with zeros to reach the required size of tokens.

\subsection{Case Study}
\begin{figure}[t]
\begin{center}
\includegraphics[width=0.6\linewidth]{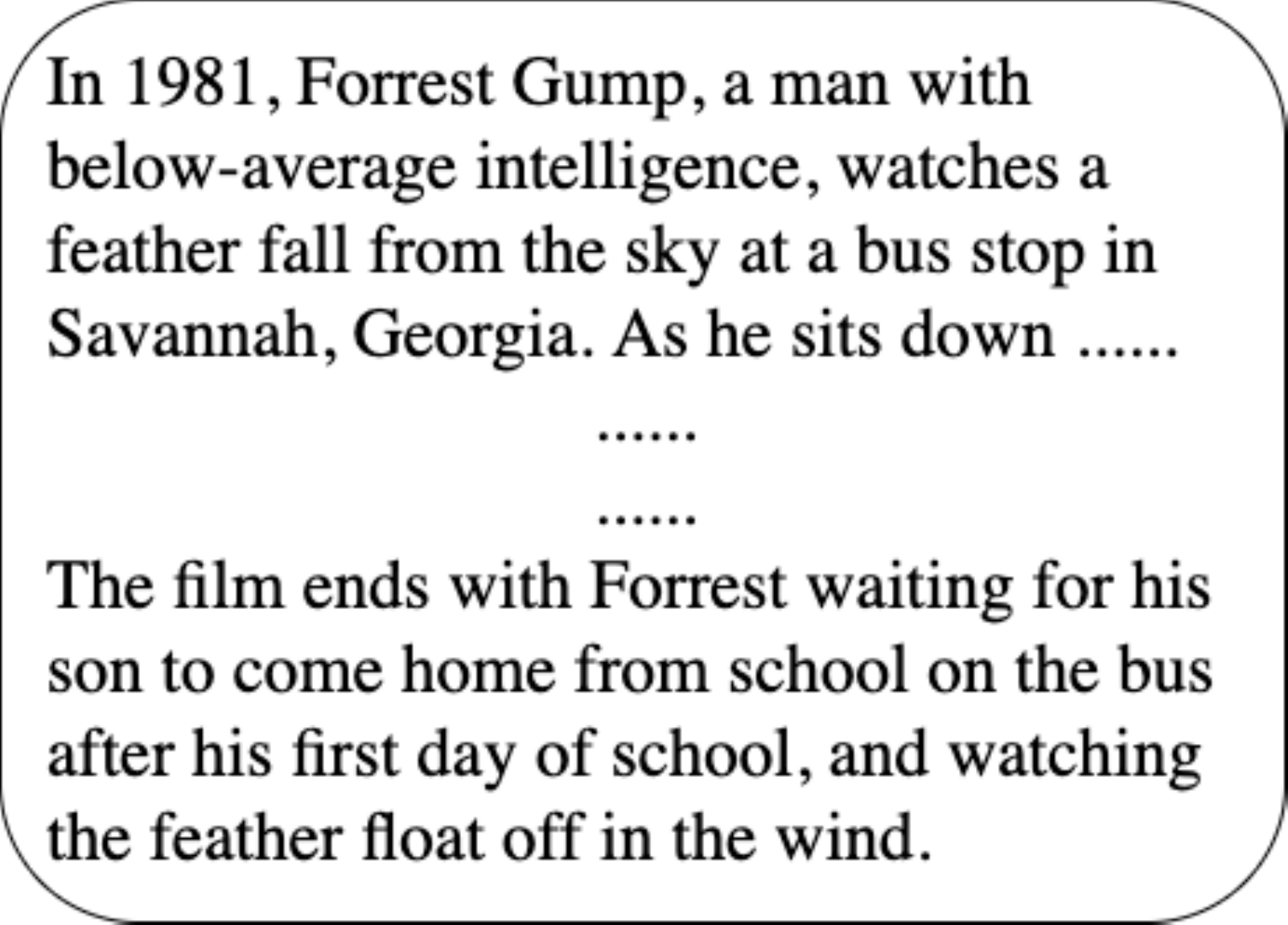}
\end{center}
\caption{Forrest Gump (1994): Plot}
\label{fig:plot}
\end{figure}

\begin{figure}
\begin{center}
\includegraphics[width=0.98\linewidth]{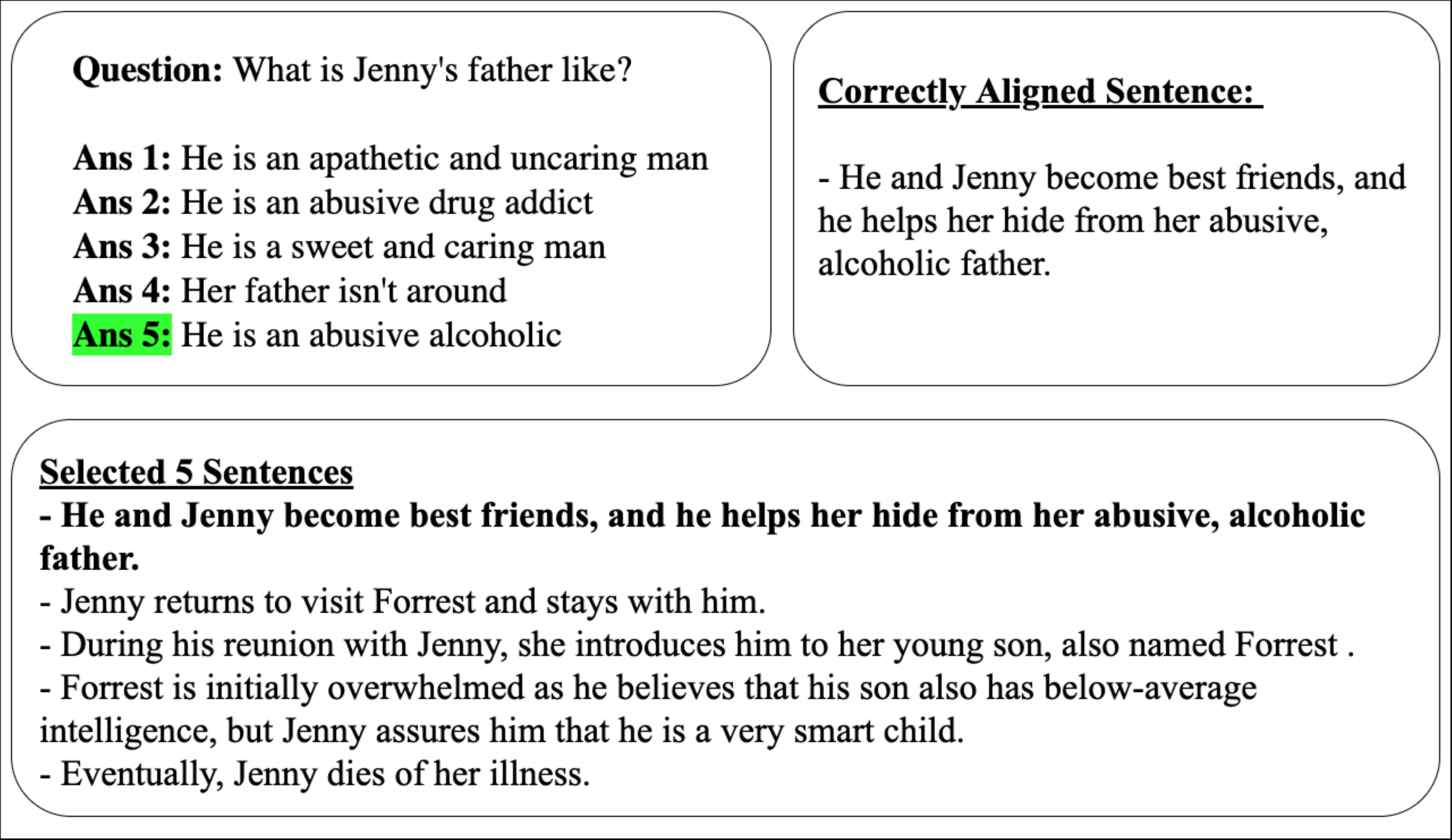}
\end{center}
\caption{Correct model prediction example}
\label{fig:correct}
\end{figure}

\begin{figure}
\begin{center}
\includegraphics[width=0.98\linewidth]{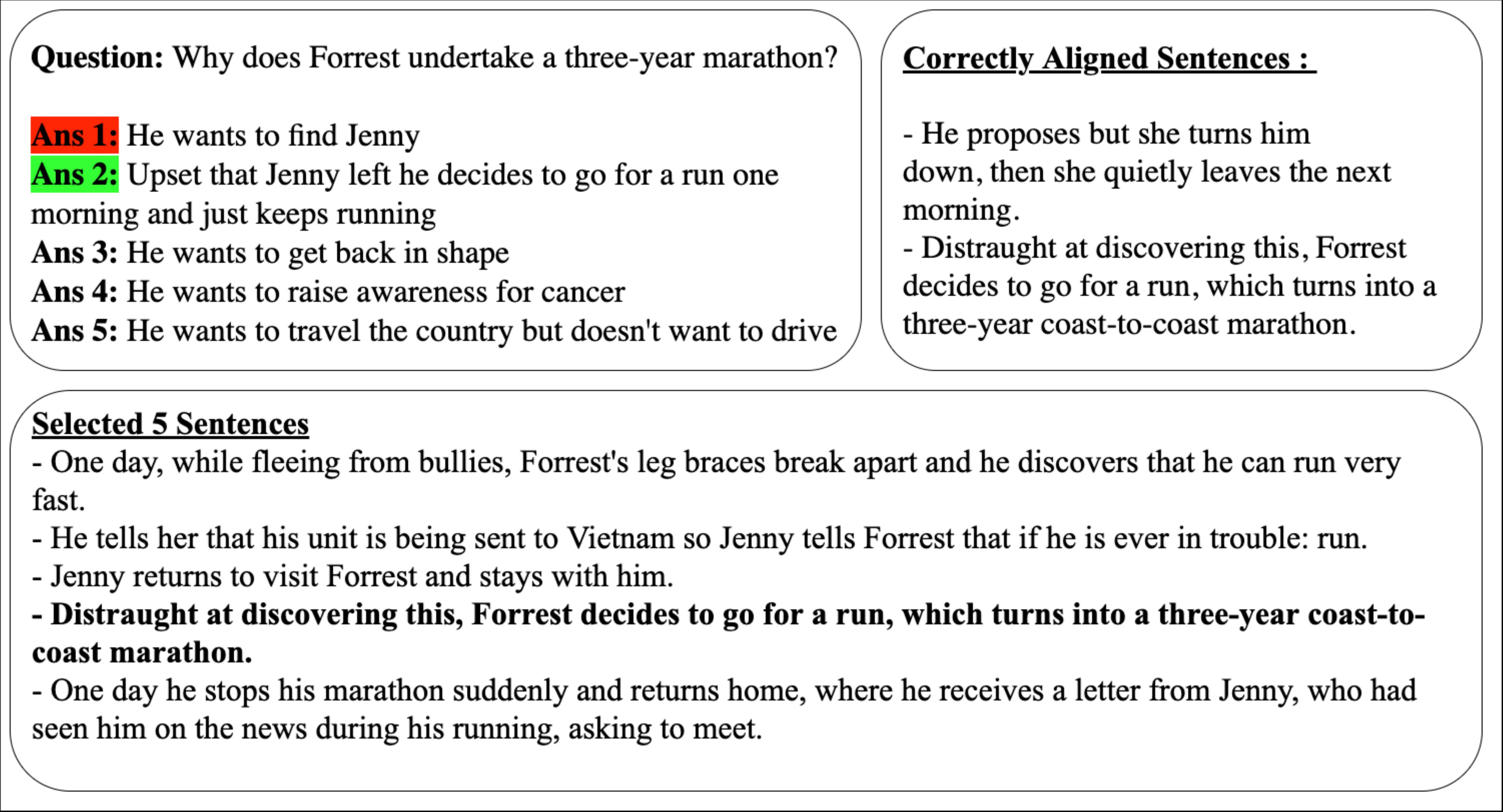}
\end{center}
\caption{Wrong model prediction example}
\label{fig:wrong}
\end{figure}

We demonstrate two cases extracted from the movie: \textit{Forrest Gump (1994)}. The movie plot is 59 sentences and a brief extract is shown in Fig. \ref{fig:plot}. Two sample questions are displayed: In Fig. \ref{fig:correct} the model selects the top 5 similar sentences to the question and in that case, the answer can be fully interpreted from one of theses sentences (highlighted in bold). Hence, after passing these sentences instead of the entire plot to BERT MCQ, it successfully selects the correct choice. 
On the other hand, the similarity model wasn't able to select the best sentences in the second example as shown in Fig. \ref{fig:wrong}. Despite finding one of the accurate sentences for this specific question, the model missed the most informative one. Therefore, the QA model subsequently failed to select the correct answer.


\section{Conclusions and Future Work}
\label{sec:conc}
In this paper, we have created an attention model based on semantic similarity to overcome \textit{BERT} limitations. In order to solve an MCQ, we begin by extracting the most relevant sentences from a large text, thereby reducing the complexity of the problem of answering MCQ question. At the time of writing this report, our latest submission is \textbf{ranked first} in the \textit{MovieQA} Challenge.\par
As a future work, we plan to extend our model to process other input signals provided by the \textit{MovieQA} dataset like subtitles. We could build more powerful models by incorporating other human like processing mechanisms such as referential relations, entailment, and answer by elimination.
Finally, migrating the code to work on TPUs with higher computational power instead of GPUs may allow us to handle larger texts avoiding sentence truncation.\par

\clearpage
\bibliographystyle{acm}
\bibliography{main}

\end{document}